\documentclass[10pt,twocolumn,letterpaper]{article}

\usepackage{cvpr}              %

\usepackage{mathtools}
\usepackage{amsmath}
\usepackage{dsfont}
\usepackage{booktabs} %
\usepackage{multirow} %

\newcommand{\TableTNTTenK}{%
\begin{table}[t]
\centering
\caption{Image metrics across several view selection metrics for the Tanks and Temples dataset after $10$K gradient steps. Two scenes are visualized, while averages are computed across the entire dataset.}
\label{fig:tnt_comparison_10K}
\setlength{\tabcolsep}{3pt}%
\scriptsize
\resizebox{\columnwidth}{!}{%
\begin{tabular}{llcccccc}
\toprule
Scene & Method & PSNR$\uparrow$ & $\Delta$PSNR & SSIM$\uparrow$ & $\Delta$SSIM & LPIPS$\downarrow$ & $\Delta$LPIPS \\
\midrule
\multirow{4}{*}{Ignatius}
& Bayes' Rays      & 16.71 & -0.24 & 0.43 & -0.12 & 0.56 & 0.24 \\
& FisherRF         & 18.01 & -0.56 & 0.59 & -0.03 & 0.31 & 0.02 \\
& Random           & 18.45 &  --   & 0.64 &  --   & 0.28 & --   \\
& \textbf{CONVERGE (Ours)} & \textbf{19.40} & \textbf{0.45} & \textbf{0.67} & \textbf{0.01} & \textbf{0.27} & \textbf{-0.01} \\
\midrule
\multirow{4}{*}{Train}
& Bayes' Rays      & 12.70 & -1.85 & 0.45 & -0.07 & 0.73 & 0.28 \\
& FisherRF         & 15.03 & -0.77 & 0.57 & -0.04 & 0.39 & 0.04 \\
& Random           & 16.37 &  --   & 0.64 &  --   & 0.32 & --   \\
& \textbf{CONVERGE (Ours)} & \textbf{17.00} & \textbf{0.65} & \textbf{0.66} & \textbf{0.02} & \textbf{0.30} & \textbf{-0.02} \\
\midrule
\multirow{4}{*}{Overall}
& Bayes' Rays      & 14.46 & -1.86 & 0.42 & -0.13 & 0.67 & 0.29 \\
& FisherRF         & 16.25 & -1.36 & 0.57 & -0.05 & 0.36 & 0.05 \\
& Random           & 18.00 &  --   & 0.65 &  --   & 0.29 & --   \\
& \textbf{CONVERGE (Ours)} & \textbf{18.34} & \textbf{0.28} & \textbf{0.66} & \textbf{0.01} & \textbf{0.28} & \textbf{-0.01} \\
\bottomrule
\end{tabular}}
\end{table}%
}

\newcommand{\TableTNTThirtyK}{%
\begin{table}[t]
\centering
\caption{Image metrics across several view selection metrics for the Tanks and Temples dataset after $30$K gradient steps. Two scenes are visualized, while averages are computed across the entire dataset.}
\label{fig:tnt_comparison_30K}
\setlength{\tabcolsep}{3pt}%
\scriptsize
\resizebox{\columnwidth}{!}{%
\begin{tabular}{llcccccc}
\toprule
Scene & Method & PSNR$\uparrow$ & $\Delta$PSNR & SSIM$\uparrow$ & $\Delta$SSIM & LPIPS$\downarrow$ & $\Delta$LPIPS \\
\midrule
\multirow{4}{*}{Ignatius}
& Bayes' Rays      & 17.10 & -2.06 & 0.46 & -0.20 & 0.45 & 0.25 \\
& FisherRF         & 20.95 & -0.28 & 0.71 & -0.03 & 0.22 & 0.02 \\
& Random           & 20.89 &  --   & 0.72 &  --   & 0.21 & --   \\
& \textbf{CONVERGE (Ours)} & \textbf{21.53} & \textbf{0.68} & \textbf{0.73} & \textbf{0.02} & \textbf{0.20} & \textbf{-0.01} \\
\midrule
\multirow{4}{*}{Train}
& Bayes' Rays      & 13.04 & -4.21 & 0.46 & -0.20 & 0.70 & 0.41 \\
& FisherRF         & 18.93 & -1.34 & 0.72 & -0.06 & 0.25 & 0.07 \\
& Random           & 19.97 &  --   & 0.76 &  --   & 0.19 & --   \\
& \textbf{CONVERGE (Ours)} & \textbf{20.30} & \textbf{0.60} & \textbf{0.77} & \textbf{0.01} & \textbf{0.18} & \textbf{-0.02} \\
\midrule
\multirow{4}{*}{Overall}
& Bayes' Rays      & 15.03 & -3.91 & 0.43 & -0.23 & 0.62 & 0.37 \\
& FisherRF         & 20.26 & -1.32 & 0.71 & -0.06 & 0.23 & 0.05 \\
& Random           & 21.12 &  --   & 0.75 &  --   & 0.20 & --   \\
& \textbf{CONVERGE (Ours)} & \textbf{21.52} & \textbf{0.41} & \textbf{0.76} & \textbf{0.01} & \textbf{0.19} & \textbf{-0.01} \\
\bottomrule
\end{tabular}}
\end{table}%
}

\newcommand{\TableMipNeRFTenK}{%
\begin{table}[t]
\centering
\caption{Image metrics across several view selection metrics for the MipNeRF360 dataset after $10$K gradient steps. Two scenes are visualized, while averages are computed across the entire dataset.}
\label{fig:mipnerf360_comparison_10K}
\setlength{\tabcolsep}{3pt}%
\scriptsize
\resizebox{\columnwidth}{!}{%
\begin{tabular}{llcccccc}
\toprule
Scene & Method & PSNR$\uparrow$ & $\Delta$PSNR & SSIM$\uparrow$ & $\Delta$SSIM & LPIPS$\downarrow$ & $\Delta$LPIPS \\
\midrule
\multirow{4}{*}{Counter}
& Bayes' Rays      & 19.74 & -2.51 & 0.61 & -0.15 & 0.55 & 0.30 \\
& FisherRF         & 22.22 & -1.66 & 0.79 & -0.04 & 0.25 & 0.04 \\
& Random           & 24.06 &  --   & 0.83 &  --   & 0.22 & --   \\
& \textbf{CONVERGE (Ours)} & \textbf{25.24} & \textbf{0.71} & \textbf{0.86} & \textbf{0.02} & \textbf{0.18} & \textbf{-0.02} \\
\midrule
\multirow{4}{*}{Garden}
& Bayes' Rays      & 17.62 & -5.32 & 0.31 & -0.31 & 0.77 & 0.51 \\
& FisherRF         & 23.73 & -0.95 & 0.72 & -0.04 & 0.21 & 0.03 \\
& Random           & 25.61 &  --   & 0.79 &  --   & 0.15 & --   \\
& \textbf{CONVERGE (Ours)} & \textbf{26.34} & \textbf{0.46} & \textbf{0.81} & \textbf{0.01} & \textbf{0.15} & \textbf{-0.01} \\
\midrule
\multirow{4}{*}{Overall}
& Bayes' Rays      & 18.07 & -2.94 & 0.44 & -0.17 & 0.68 & 0.35 \\
& FisherRF         & 20.99 & -1.40 & 0.67 & -0.04 & 0.29 & 0.05 \\
& Random           & 22.89 &  --   & 0.72 &  --   & 0.23 & --   \\
& \textbf{CONVERGE (Ours)} & \textbf{23.36} & \textbf{0.29} & \textbf{0.72} & \textbf{0.00} & \textbf{0.22} & \textbf{-0.01} \\
\bottomrule
\end{tabular}}
\end{table}%
}

\newcommand{\TableMipNeRFThirtyK}{%
\begin{table}[t]
\centering
\caption{Image metrics across several view selection metrics for the MipNeRF360 dataset after $30$K gradient steps. Two scenes are visualized, while averages are computed across the entire dataset.}
\label{fig:mipnerf360_comparison_30K}
\setlength{\tabcolsep}{3pt}%
\scriptsize
\resizebox{\columnwidth}{!}{%
\begin{tabular}{llcccccc}
\toprule
Scene & Method & PSNR$\uparrow$ & $\Delta$PSNR & SSIM$\uparrow$ & $\Delta$SSIM & LPIPS$\downarrow$ & $\Delta$LPIPS \\
\midrule
\multirow{4}{*}{Counter}
& Bayes' Rays      & 20.39 & -4.90 & 0.62 & -0.22 & 0.48 & 0.32 \\
& FisherRF         & 26.35 & -1.73 & 0.87 & -0.04 & 0.17 & 0.04 \\
& Random           & 27.51 &  --   & 0.89 &  --   & 0.15 & --   \\
& \textbf{CONVERGE (Ours)} & \textbf{27.96} & \textbf{0.79} & \textbf{0.90} & \textbf{0.02} & \textbf{0.14} & \textbf{-0.02} \\
\midrule
\multirow{4}{*}{Garden}
& Bayes' Rays      & 19.28 & -7.33 & 0.33 & -0.44 & 0.75 & 0.59 \\
& FisherRF         & 27.77 & -1.01 & 0.85 & -0.04 & 0.12 & 0.03 \\
& Random           & 27.85 &  0.00 & 0.85 &  0.00 & 0.11 & 0.00 \\
& \textbf{CONVERGE (Ours)} & \textbf{28.06} & \textbf{0.51} & \textbf{0.86} & \textbf{0.01} & \textbf{0.10} & \textbf{-0.01} \\
\midrule
\multirow{4}{*}{Overall}
& Bayes' Rays      & 18.81 & -4.92 & 0.45 & -0.25 & 0.63 & 0.41 \\
& FisherRF         & 24.92 & -1.35 & 0.78 & -0.03 & 0.19 & 0.04 \\
& Random           & 25.57 &  0.00 & 0.78 &  0.00 & 0.17 & 0.00 \\
& \textbf{CONVERGE (Ours)} & \textbf{25.78} & \textbf{0.35} & \textbf{0.78} & \textbf{0.00} & \textbf{0.17} & \textbf{-0.01} \\
\bottomrule
\end{tabular}}
\end{table}%
}

\newcommand{\TableCapturesTenK}{%
\begin{table}[t]
\centering
\caption{Image metrics across several view selection metrics for the custom dataset after $10$K gradient steps. Two scenes are visualized, while averages are computed across the entire dataset.}
\label{fig:custom_comparison_10K}
\setlength{\tabcolsep}{3pt}%
\scriptsize
\resizebox{\columnwidth}{!}{%
\begin{tabular}{llcccccc}
\toprule
Scene & Method & PSNR$\uparrow$ & $\Delta$PSNR & SSIM$\uparrow$ & $\Delta$SSIM & LPIPS$\downarrow$ & $\Delta$LPIPS \\
\midrule
\multirow{4}{*}{Shiny}
& Bayes' Rays      & 13.08 & -1.43 & 0.31 & -0.09 & 0.88 & 0.29 \\
& FisherRF         & 13.69 & -1.35 & 0.38 & -0.05 & 0.58 & 0.07 \\
& Random           & 15.47 &  --   & 0.46 &  --   & 0.47 & --   \\
& \textbf{CONVERGE (Ours)} & \textbf{16.49} & \textbf{0.76} & \textbf{0.49} & \textbf{0.02} & \textbf{0.43} & \textbf{-0.03} \\
\midrule
\multirow{4}{*}{Space}
& Bayes' Rays      & 11.70 & -1.74 & 0.20 & -0.13 & 0.85 & 0.34 \\
& FisherRF         & 13.94 & -0.55 & 0.39 & -0.03 & 0.46 & 0.03 \\
& Random           & 14.88 &  --   & 0.45 &  --   & 0.41 & --   \\
& \textbf{CONVERGE (Ours)} & \textbf{15.66} & \textbf{0.78} & \textbf{0.48} & \textbf{0.03} & \textbf{0.36} & \textbf{-0.03} \\
\midrule
\multirow{4}{*}{Overall}
& Bayes' Rays      & 13.20 & -1.62 & 0.27 & -0.09 & 0.84 & 0.35 \\
& FisherRF         & 14.58 & -1.02 & 0.38 & -0.05 & 0.48 & 0.06 \\
& Random           & 16.11 &  --   & 0.46 &  --   & 0.38 & --   \\
& \textbf{CONVERGE (Ours)} & \textbf{16.94} & \textbf{0.68} & \textbf{0.50} & \textbf{0.02} & \textbf{0.35} & \textbf{-0.03} \\
\bottomrule
\end{tabular}}
\end{table}%
}

\newcommand{\TableCapturesThirtyK}{%
\begin{table}[t]
\centering
\caption{Image metrics across several view selection metrics for the custom dataset after $30$K gradient steps. Two scenes are visualized, while averages are computed across the entire dataset.}
\label{fig:custom_comparison_30K}
\setlength{\tabcolsep}{3pt}%
\scriptsize
\resizebox{\columnwidth}{!}{%
\begin{tabular}{llcccccc}
\toprule
Scene & Method & PSNR$\uparrow$ & $\Delta$PSNR & SSIM$\uparrow$ & $\Delta$SSIM & LPIPS$\downarrow$ & $\Delta$LPIPS \\
\midrule
\multirow{4}{*}{Shiny}
& Bayes' Rays      & 13.38 & -3.08 & 0.30 & -0.17 & 0.86 & 0.42 \\
& FisherRF         & 16.97 & -1.59 & 0.52 & -0.06 & 0.41 & 0.09 \\
& Random           & 18.27 &  --   & 0.56 &  --   & 0.33 & --   \\
& \textbf{CONVERGE (Ours)} & \textbf{18.41} & \textbf{0.59} & \textbf{0.58} & \textbf{0.02} & \textbf{0.31} & \textbf{-0.03} \\
\midrule
\multirow{4}{*}{Space}
& Bayes' Rays      & 11.76 & -3.85 & 0.19 & -0.27 & 0.83 & 0.46 \\
& FisherRF         & 18.16 & -0.28 & 0.58 & -0.02 & 0.29 & 0.03 \\
& Random           & 17.85 &  --   & 0.58 &  --   & 0.27 & --   \\
& \textbf{CONVERGE (Ours)} & \textbf{18.24} & \textbf{0.69} & \textbf{0.59} & \textbf{0.02} & \textbf{0.25} & \textbf{-0.03} \\
\midrule
\multirow{4}{*}{Overall}
& Bayes' Rays      & 13.40 & -3.42 & 0.27 & -0.21 & 0.82 & 0.46 \\
& FisherRF         & 17.94 & -1.14 & 0.54 & -0.06 & 0.33 & 0.08 \\
& Random           & 18.71 &  --   & 0.58 &  --   & 0.27 & --   \\
& \textbf{CONVERGE (Ours)} & \textbf{19.05} & \textbf{0.60} & \textbf{0.60} & \textbf{0.02} & \textbf{0.25} & \textbf{-0.02} \\
\bottomrule
\end{tabular}}
\end{table}%
}

\newcommand{\TableSummaryThirtyK}{%
\begin{table}[t]
\centering
\caption{Image metrics of different view-selection methods at 30K steps across different datasets.}
\label{tab:summary_30k_structured}
\setlength{\tabcolsep}{3pt}%
\scriptsize
\resizebox{\columnwidth}{!}{%
\begin{tabular}{llccc}
\toprule
Dataset & Method & PSNR$\uparrow$ & SSIM$\uparrow$ & LPIPS$\downarrow$ \\
\midrule
\multirow{4}{*}{Tanks \& Temples}
& Bayes' Rays & 15.03 & 0.43 & 0.62 \\
& FisherRF & 20.26 & 0.71 & 0.23 \\
& Random & 21.12 & 0.75 & 0.20 \\
& \textbf{CONVERGE (Ours)} & \textbf{21.52} & \textbf{0.76} & \textbf{0.19} \\
\midrule
\multirow{4}{*}{Mip-NeRF}
& Bayes' Rays & 18.81 & 0.45 & 0.63 \\
& FisherRF & 24.92 & 0.78 & 0.19 \\
& Random & 25.57 & 0.78 & 0.17 \\
& \textbf{CONVERGE (Ours)} & \textbf{25.78} & \textbf{0.78} & \textbf{0.17} \\
\midrule
\multirow{4}{*}{Captures}
& Bayes' Rays & 13.40 & 0.27 & 0.82 \\
& FisherRF & 17.94 & 0.54 & 0.33 \\
& Random & 18.71 & 0.58 & 0.27 \\
& \textbf{CONVERGE (Ours)} & \textbf{19.05} & \textbf{0.60} & \textbf{0.25} \\
\midrule
\multirow{4}{*}{Overall}
& Bayes' Rays & 15.75 & 0.38 & 0.69 \\
& FisherRF & 21.04 & 0.68 & 0.25 \\
& Random & 21.80 & 0.70 & 0.21 \\
& \textbf{CONVERGE (Ours)} & \textbf{22.12} & \textbf{0.71} & \textbf{0.20} \\
\bottomrule
\end{tabular}}%
\end{table}%
}
\newcommand{\TableEmbodiedSparseAblationThirtyK}{%
\begin{table}[t]
\centering
\caption{Image metrics across different view-selection settings at 30K steps,  averaged over all scenes. Splatfacto, with access to all views, serves as the infeasible upper-bound.}
\label{tab:ablation_30k_settings}
\setlength{\tabcolsep}{3pt}%
\scriptsize
\resizebox{\columnwidth}{!}{%
\begin{tabular}{llccc}
\toprule
Setting & Method & PSNR $\uparrow$ & SSIM $\uparrow$ & LPIPS $\downarrow$ \\
\midrule
All & Splatfacto & 24.83 & 0.79 & 0.16 \\
\midrule
\multirow{3}{*}{Embodied}
    & Random & 22.48 & 0.71 & 0.23 \\
    & Fisher-RF & 22.27 & 0.70 & 0.24 \\
    & \textbf{CONVERGE (Ours)}  & \textbf{23.21} & \textbf{0.73} & \textbf{0.20} \\
\midrule
\multirow{3}{*}{Sparse}
    & Random & \textbf{22.81} & \textbf{0.72} & 0.21 \\
    & Fisher-RF & 21.74 & 0.68 & 0.26 \\
    & \textbf{CONVERGE (Ours)} & 22.80 & 0.71 & \textbf{0.21} \\
\midrule
\multirow{3}{*}{Embodied + Sparse}
    & Random & 20.89 & 0.65 & 0.32 \\
    & Fisher-RF & 21.24 & 0.66 & 0.31 \\
    & \textbf{CONVERGE (Ours)}  & \textbf{22.39} & \textbf{0.70} & \textbf{0.24} \\
\bottomrule
\end{tabular}}%
\end{table}%
}

\newcommand{\TableEmbodiedAblationThirtyK}{%
\begin{table}[t]
\centering
\caption{Image metrics across different view-selection methods at 30K steps,  averaged over all scenes.}
\label{tab:ablation_30k_grad}
\setlength{\tabcolsep}{3pt}%
\scriptsize
\resizebox{\columnwidth}{!}{%
\begin{tabular}{llccc}
\toprule
Setting & Method & PSNR $\uparrow$ & SSIM $\uparrow$ & LPIPS $\downarrow$ \\
\midrule
\multirow{3}{*}{Embodied + Sparse}
    & \Cref{eq:view_metric_linearized_beta} & 21.95 & 0.695 & 0.251 \\
    & Gradient & 21.77 & 0.701 & 0.275 \\
    & Coverage & \textbf{22.39} & \textbf{0.707} & \textbf{0.244} \\
\bottomrule
\end{tabular}}%
\end{table}%
}

\definecolor{cvprblue}{rgb}{0.21,0.49,0.74}
\usepackage[pagebackref,breaklinks,colorlinks,allcolors=cvprblue]{hyperref}
\usepackage{caption}
\usepackage{graphicx}
\usepackage{capt-of}

\title{Coverage Optimization for Camera View Selection}

\author{%
\textbf{Timothy Chen}\textsuperscript{*}\hspace{0.8em}\textbf{Adam Dai}\textsuperscript{*}\hspace{0.8em}\textbf{Maximilian Adang}\hspace{0.8em}\textbf{Grace Gao}\hspace{0.8em}\textbf{Mac Schwager}\\[0.5em]
Stanford University\\
}

\begin{document}
\twocolumn[{%
\renewcommand\twocolumn[1][]{#1}%
\maketitle

\begin{center}
    \centering
   \captionsetup{type=figure}
    \includegraphics[width=\textwidth]{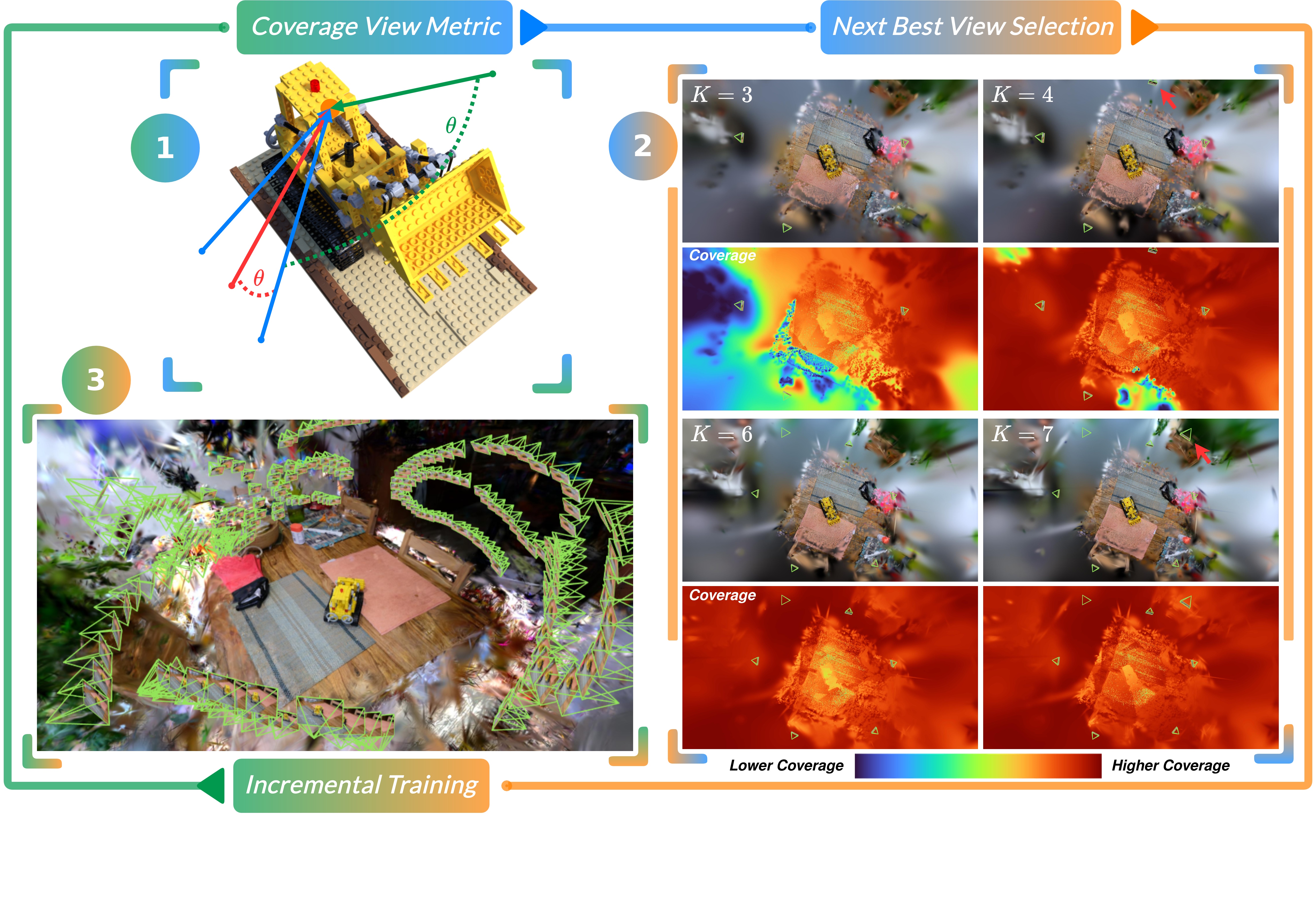}
    \vspace{-5em}
    \captionof{figure}{\textbf{COVER} is a simple, performant, and from-first-principles view selection metric that can be batch-queried in real-time and rendered into an image for visualization. The view metric measures the difference between perspective candidate cameras and cameras in the training dataset. Candidate viewpoints that cover large parts of the scene and are under-covered by the training cameras are selected and added to the training dataset to improve the quality of the $3$D scene reconstruction.}
\end{center}%
}]
\begingroup
\renewcommand{\thefootnote}{\fnsymbol{footnote}}%
\footnotetext[1]{Equal contribution.}%
\endgroup

\begin{abstract}
What makes a good viewpoint? 
The quality of the data used to learn 3D reconstructions is crucial for enabling efficient and accurate scene modeling. 
We study the active view selection problem and develop a principled analysis that yields a simple and interpretable criterion for selecting informative camera poses. 
Our key insight is that informative views can be obtained by minimizing a tractable approximation of the Fisher Information Gain, which reduces to favoring viewpoints that cover geometry that has been insufficiently observed by past cameras. 
This leads to a lightweight coverage-based view selection metric that avoids expensive transmittance estimation and is robust to noise and training dynamics. We call this metric COVER (Camera Optimization for View Exploration and Reconstruction).
We integrate our method into the Nerfstudio framework and evaluate it on real datasets within fixed and embodied data acquisition scenarios. 
Across multiple datasets and radiance-field baselines, our method consistently improves reconstruction quality compared to state-of-the-art active view selection methods. Additional visualizations and our Nerfstudio package can be found at our \href{https://chengine.github.io/nbv_gym/}{webpage}.

\end{abstract}

\section{Introduction}

Progress in photorealistic scene reconstruction has advanced rapidly, enabling the geometry and appearance of real environments to be recovered in real time. 
However, much of this progress relies on access to high-quality training views. 
Even with ideal sensing, a fundamental question about observability remains: where should new viewpoints be placed to maximize the quality of the reconstructed scene? Although this problem is ill-posed due to the unavailability of the ground-truth geometry, we observe that human-captured datasets naturally yield well-constructed scenes. 

A number of heuristic strategies exist for acquiring new views autonomously, yet many perform only marginally better than random sampling in specific instances. 
More principled techniques based on information theory—such as FisherRF or uncertainty quantification in NeRFs—offer deeper mathematical grounding but are computationally expensive, sensitive to training noise, are conditioned on non-stationary quantities that can change rapidly during optimization, and require custom CUDA kernels.

Despite the apparent complexity of the problem, humans routinely capture informative views with little effort, suggesting the existence of simple rules that are not fully explored mathematically. 
We revisit this problem from first principles and show that, under a natural approximation to the Fisher Information Gain, an informative next view is one that observes geometry poorly covered by previous viewpoints. 
This perspective naturally leads to a coverage-based view selection metric that is computationally lightweight, highly interpretable, and compatible with modern radiance field pipelines.

In this work, we make the following contributions:
(1) We derive a tractable approximation to the Fisher Information Gain that identifies the primitives whose parameters are not fully constrained by existing training views.
(2) We show that, under this approximation, selecting informative viewpoints reduces to minimizing a simple coverage metric that depends only on per-primitive visibility, not noisy and non-stationary quantities like transmittance.
(3) We integrate this metric, as well as competing baselines, into the Nerfstudio~\cite{tancik2023nerfstudio} training pipeline and evaluate them across 15 real human-captured scenes in a fixed dataset and embodied data acquisition scenario. Our coverage-optimized view selection consistently outperforms state-of-the-art and randomized baselines in reconstruction quality, despite the already well-covered nature of the datasets. The gap is extended for 
embodied data acquisition scenarios, suggesting natural compatibility between COVER and robot deployments in the wild.

\section{Related Work}
\label{Sec:RelatedWork}

The problem of selecting camera views that improve reconstruction or render quality has gained attention in radiance field literature. Pan et al.~\cite{pan2022activenerf} incorporate model uncertainty into the resource constrained view-selection problem with ActiveNeRF by expanding the training dataset to maximize information gain. Yan and Liu et al.~\cite{yan2023active_implicit} build an implicit volumetric occupancy field and extract its entropy as a measure of novel view information gain, using a sampling-based planner to accelerate object-level reconstruction through next-best-view planning. Goli et al.~\cite{goli2024bayes} provide a post-hoc uncertainty measure which may be used to guide informative view acquisition for 3D reconstruction. Xie and Zhang et al. develop S-NeRF~\cite{xie2023s} to complement this by providing a structured, part-aware uncertainty representation within the NeRF itself. Lin and Yi use ensemble-based NeRFs~\cite{lin2022active} to estimate epistemic uncertainty through variance across multiple independently trained networks in a model-agnostic alternative. Most similar to our work, Jiang et al.~\cite{jiang2023fisherrf} formulate FisherRF, which quantifies the information gain of a candidate pose by computing the Fisher Information of the radiance field---offering a metric extensible to simultaneous robot motion planning and 3D reconstruction. Strong et al.~\cite{strong2025next} extend FisherRF to formulate a depth-based uncertainty metric for next-best-view selection. Complementary approaches assess viewpoint coverage as a primary metric for improving reconstruction quality: Xiao et al.~\cite{xiao2024nerf} demonstrate that uniform coverage of objects outperforms complex uncertainty metrics, while Xue et al.~\cite{xue2024neural} incorporate visibility-driven uncertainty into robotic next-best-view selection. Li et al.~\cite{li2025activesplat} balance exploration efficiency with complete coverage using hybrid map representations, and Xu et al.~\cite{xu2025hgs} explicitly assess coverage of unexplored areas by integrating unknown voxels into the rendering process with HGS-Planner. Tao et al.~\cite{tao2025rt} use the changing magnitude of parameters of a 3DGS during reconstruction to actively plan onboard a robot in RT-Guide. Nagami et al.~\cite{nagami2025vista} propose VISTA, a semantic exploration strategy for robots using online Gaussian splatting and a geometric information gain metric to guide robot motion towards informative views. While prior work treats optimizing for information gain and spatial coverage as separate objectives, no existing method provides a unified framework that shows their equivalence. Our approach addresses this gap by formulating next-best-view selection that explicitly reconciles information gain and viewpoint coverage in a principled way.

\section{Preliminaries}

\subsection{Radiance Fields}
Radiance fields were first formulated by Williams and Max ~\cite{max1992dvolumedensity, max1995opticalmodel} and popularized using neural networks and modern GPUs in NeRF \cite{mildenhall2021nerf}. A radiance field is composed of two distinct fields. The geometry of a 3D scene is softly modeled using a density field $\rho(x): \mathbb{R}^3 \to \mathbb{R}_+$. The texture and specularities are modeled using a radiance field $c(x, d): \mathbb{R}^3 \times \mathbb{S}^2 \to [0, 1]^3$, conditioned on a 3D position $x$ and a viewing direction $d$. An RGB image can be rendered from these two fields on a per-pixel level (conditioned on a ray defined by origin $x_0$ and direction $d$) using a discrete form of the radiance field rendering equation
\begin{align}
    \label{eq:radiance_render}
    C(x_0, d) &= \sum_i^N \underbrace{T_i(t_{1:i}; x_0, d) \sigma_i(t_i; x_0, d)}_{w_i(t_i; x_0, d)} c_i(t_i; x_0, d),\\
    \text{where } T_i &= \prod_{j=1}^{i-1} (1 - \sigma_j(t_j; x_0, d) ).
\end{align}
Other attributes beyond color can be easily rendered using \cref{eq:radiance_render}, like depth or semantic embeddings \cite{shen2023F3RM,shorinwa2024splat}.

\subsection{Gaussian Splatting}
While our proceeding analysis is general to any radiance field representation, we introduce a popular state-of-the-art representation that we use in our implementation of view selection. Gaussian Splatting (3DGS) \cite{kerbl2023splatting} is an efficient extension of NeRFs \cite{mildenhall2021nerf} , encoding an occupancy and  radiance. Instead of parameterizing these fields with a neural network, the authors recognized that these fields are spatially sparse, choosing to model the occupied space with ellipsoidal primitives with occupancy and radiance attributes. Simply, 3DGS is an augmented point cloud with $N$ points $\mathcal{G} = \{{G}_i = (\mu, \Sigma, c, \sigma) \}_i^N$, where each point contains information about its mean (location) $\mu \in \mathbb{R}^3$, covariance (extent) $\Sigma \in S_3^{++}$, radiance $c \in [0, 1]^3$, and occupancy $\sigma \in [0, 1]$. Unlike NeRF's method of rendering images using volumetric ray-tracing, Gaussian Splatting projects 3D Gaussians onto the image plane (i.e. rasterization). In this way, 3DGS is more efficient and does not allocate resources to model empty regions of $3$D space. Consequently, Gaussian Splatting demonstrates comparable if not better photorealism than NeRFs, while exhibiting faster training and rendering times. 

\section{Method}
\label{Sec:Method}

Our derivation of an interpretable and tractable information gain metric proceeds in three steps: (1) expressing Fisher Information Gain as a quadratic form over transmittance patterns; (2) extending this metric to a view-direction-aware formulation; and (3) relaxing this form to a coverage-based surrogate.

\subsection{Formulating a Gain Metric}
Given a scene with $P$ primitives and a dataset $\mathcal{D} = \{(x_0^k, d^k, C^k)\}_{k=1}^{K  Z}$ with $K$ cameras and $Z$ pixels per camera, the regression of the primitive attributes can be formulated as a least squares problem

\begin{equation}
    \label{eq:radiance_least_squares}
    \min_{c} \lVert W^{(K)} c - C\rVert_2^2,
\end{equation}
where $c\in \mathbb{R}^P$ are the per-primitive attributes and $W^{(K)} \in \mathbb{R}^{K Z\times P}$, $W^{(K)} \geq 0$ is the weight matrix containing the termination probabilities (or equivalently the transmittance pattern) $w_i$ for all primitives and all $KZ$ observations. This regression problem can also be immediately extended to multi-channel attributes like color (RGB) channels by minimizing each color channel separately. Although the vector of termination probabilities need not abide by a norm constraint (except that its elements implicitly sum to no more than 1 and live in the non-negative orthant), we assume the rows of $W^{(K)}$ are unit-norm with $C_k$ properly scaled. Normalization of the data matrix is common in linear regression and aids in interpretability of the result. 

How certain are we about the optimal value $c^*$? A paradigm has been set to use the Fisher Information as a metric to gauge the uncertainty and sensitivity of the optimal solution $c^*$. Specifically, the Fisher Information as a scalar is typically represented as the log-determinant of the Gram matrix
\begin{equation}
    \label{eq:fisher_info}
    F(W) = \log \lvert \underbrace{W^T W}_G \rvert,
\end{equation}
where we assume $G$ is full rank. 

In the active view selection problem, we ask: how does the uncertainty change by adding new views? Similarly, if we add one new observation to the regression problem, we perform a rank-one update to the Gram matrix and update the Fisher Information accordingly
\begin{equation}
    \label{eq:fisher_info_update}
    F(W^{(K+1)}) = \log \lvert G + w w^T \rvert,
\end{equation}
where $w$ is the termination probabilities vector for the new observation.
The Fisher Information \emph{Gain} (FIG) is simply the difference
\begin{equation}
    \label{eq:fisher_info_gain}
    \begin{split}
    \text{FIG}(w; W) &= \log\lvert G + w w^T \rvert -  \log \lvert G \rvert\\
    &= \log (1 + w^T G^{-1} w),
    \end{split}
\end{equation}
where $\lvert G + w w^T \rvert = \lvert G \rvert (1 + w^T G^{-1} w)$ by the \emph{matrix determinant lemma}. Note that maximizing the FIG is equivalent to maximizing the quadratic term. Under the assumption that all rows of $W^{(K)}$ and $W^{(K+1)}$ are of unit norm, which implies $w$ is also unit norm, then
\begin{equation}
    \label{eq:max_min_equivalence}
    \underset{\lVert w \rVert_2 = 1}{\arg\max} \; w^T G^{-1} w = \underset{\lVert w \rVert_2 = 1}{\arg\min} \; w^TG w.
\end{equation}
The proof of \cref{eq:max_min_equivalence} is automatic by \emph{Rayleigh quotients}. The following versions of the min-norm problems are also equivalent
\begin{equation}
    \label{eq:min-norm}
    \underset{w \in \mathcal{W}}{\arg\min} \lVert W^{(K)} w\rVert_2^2 = \underset{w \in \mathcal{W}}{\arg\min} \lVert W^{(K)} w\rVert_2,
\end{equation}
for arbitrary sets $\mathcal{W}$. 

\Cref{eq:min-norm} is very interpretable, as we desire $w$, the \emph{candidate} transmittance pattern, to be orthogonal to all the \emph{observed} transmittance patterns in order to maximize gained information on the per-primitive colors, restricting their uncertainty and pushing them toward stable unique values. 

However, note that with additional constraints, \cref{eq:max_min_equivalence} is no longer equality. Rather, we have the \emph{tight} bound by Cauchy-Schwarz
\begin{equation}
    \label{eq:fig_lower_bound}
    w^T G^{-1} w \geq \frac{1}{w^T G w},
\end{equation}
with equality when $w$ is an eigenvector of $G$. Regardless, minimizing \cref{eq:min-norm} subject to additional constraints on $w$ still applies upward pressure to the FIG. 

\subsection{Tractable Metric}
Storing $W^{(K)}$ is intractable, as the matrix has as many rows as the number of pixels in the training dataset and as many columns as the number of primitives, and continues to grow as we take more active view selection steps. Instead, we show that a proxy minimizer can be computed by simply storing a single scalar value per primitive, which is updated every time we take a step. 

We show that the following is true
\begin{equation}
    \label{eq:min_norm_sum_equals_sum_norm}
    \underset{w \in \mathbb{S}^{P-1}_+}{\arg\min} \lVert \sum_i^P W^{(K)}_{:, i} w_i\rVert_2 = \underset{w \in \mathbb{S}^{P-1}_+}{\arg\min} \sum_i^P  w_i \lVert W^{(K)}_{:, i}\rVert_2,
\end{equation}
where $\mathbb{S}^{P-1}_+ \coloneq \{w \in \mathbb{R}^P_+ | \lVert w \rVert_2 = 1 \}$ and $W_{:, i}$ are columns of $W$ in \Cref{app:proof1}.

Why is computing the RHS objective (\ref{eq:min_norm_sum_equals_sum_norm}) more tractable? Rather than storing all incident transmittance patterns over all observations per-primitive (LHS (\ref{eq:min_norm_sum_equals_sum_norm})), the RHS (\ref{eq:min_norm_sum_equals_sum_norm}) only requires storing the running norm of the incident transmittance patterns $\lVert W^{(K+1)}_{:, i} \rVert_2^2 = \lVert W^{(K)}_{:, i} \rVert_2^2 + w_i^2$. 

Moreover, notice that a linear combination of per-primitive attributes using the transmittance weights is precisely the \emph{rendering} of a pixel and mirrors the radiance rendering equation (\ref{eq:radiance_render}). As a result, using the metric 
\begin{equation}
    \label{eq:info_gain_render}
    \mathcal{I}^{(K+1)}_{\text{trans}} (x_0, d) = \sum_i^P w_i(x_0, d) \lVert W^{(K)}_{:, i} \rVert_2,
\end{equation}
is advantageous in many ways. First, the metric is computationally efficient to compute and store, and only requires appending an additional channel to the color rendering routine. Secondly, it can be visualized as an \emph{image}, making the metric interpretable and user-friendly. Finally, we have shown tight correspondence between minimizing Eq. (\ref{eq:info_gain_render}) and maximizing the Fisher Information Gain.

\begin{figure*}[th]
    \centering
    \includegraphics[width=\textwidth]{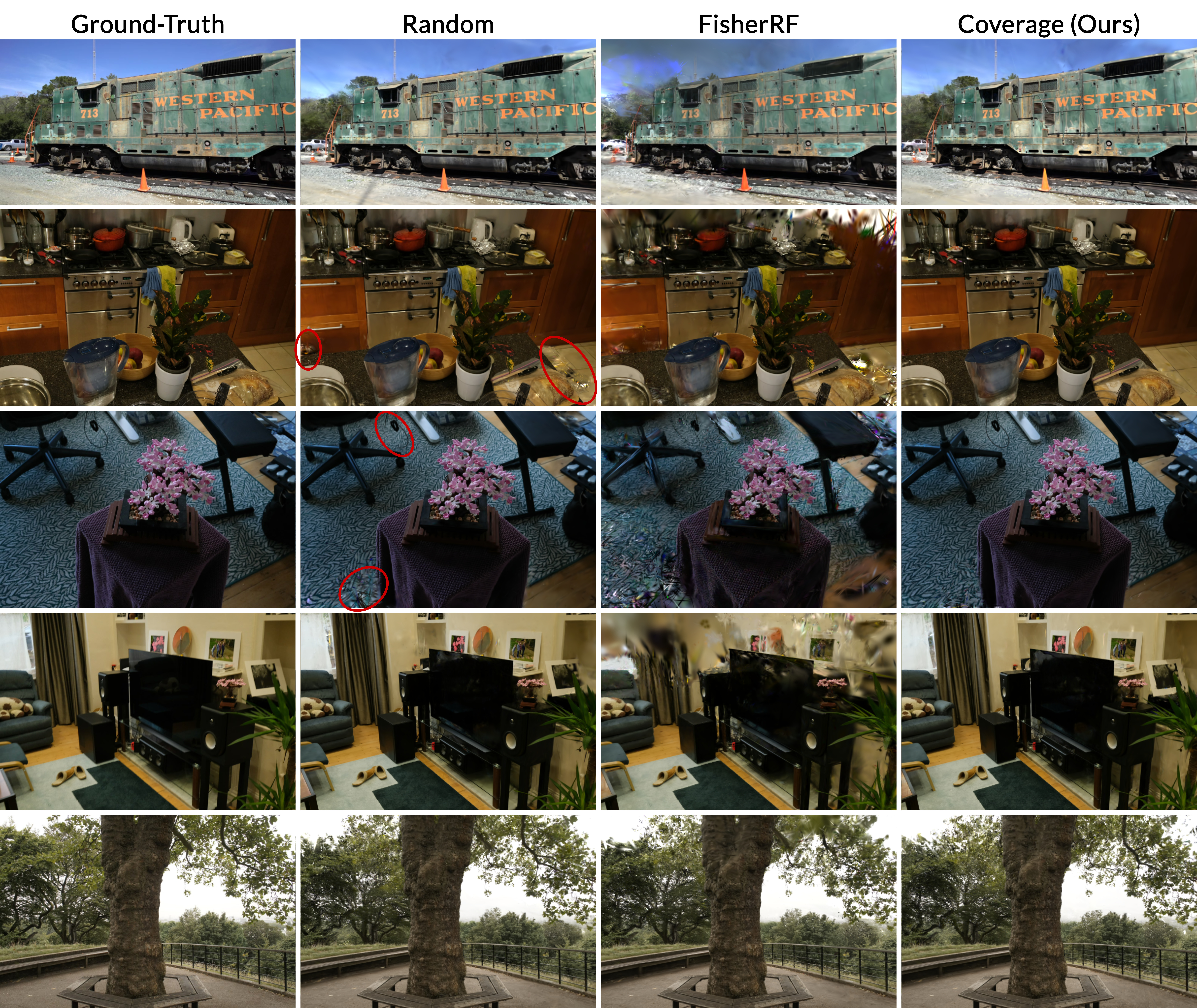}
    \caption{Renders of $3$DGS models trained under different view selection metrics. Images are from the evaluation set and not seen during training. We find that our coverage metric is at least as good as random and superior to FisherRF. We highlight lower coverage regions in the random baseline in \textcolor{red}{red}. Additional visual comparisons and visualization of the view selection process can be found on our \href{https://chengine.github.io/nbv_gym/}{webpage} and in \Cref{app:more_viz}.}
    \label{fig:evalset_viz}
\end{figure*}

\subsection{Extension to View Directions}
The previous analysis is view-direction independent, which applies to \emph{matte} attributes like matte colors or attributes solely dependent on position (e.g. semantic embeddings). In order to extend our analysis to finding optimal viewing directions rather than simply transmittance patterns, we assume a general per-primitive color model of the form

\begin{equation}
    \label{eq:color_model}
    c^i(d) = \beta^i (d) r^i,
\end{equation}
where $\beta^i \in \mathbb{R}^L_+$ is the primitive $i$'s vector of weights for patches distributed on $\mathbb{S}^2$, whose weight is determined by a spherical radial kernel centered at $d$. We assume that, like $w$, the vector of weights $\beta^i$ lives in $\mathbb{S}_+^{L-1}$. $r^i \in [0, 1]^L$ is the corresponding radiances associated with the patches, forming a color field on the unit sphere. Essentially, the color when viewed at $d$ is some weighted average of the color field, with the weights decaying away from $d$. Examples of suitable kernels are the spherical Gaussian kernel. This assumption is not a limiting one, as spherical harmonic coefficients can be optimally extracted from the color field using linear regression. 

We  extend the color regression problem (\ref{eq:radiance_least_squares}) to color field regression
\begin{align}
    \label{eq:color_field_least_squares}
    &\min_r \lVert \Tilde{W}^{(K)} r - C \rVert_2^2,\\
    &\text{and} \; \; \Tilde{W}^{(K)} = W  \; [\underbrace{\text{blkdiag}(\beta^1, ..., \beta^P)}_B],
\end{align}
where the new design matrix is the matrix product between the view-independent weight matrix and a view-dependent block-diagonal matrix $B \in \mathbb{R}_+^{P \times P L}$ of the per-primitive vector patch weights. Since $\beta^i \in \mathbb{S}_+^{L-1}$ and $w \in \mathbb{S}_+^{P-1}$,
the new observation to be added to $\Tilde{W}^{(K)}$ satisfies 
\begin{align}
\label{eq:w_space}
    \Tilde{w} &= w^T \text{blkdiag}(\beta^1, ..., \beta^P) \;   \implies  \; \lVert \Tilde{w} \rVert_2^2 = 1.
\end{align}
Our analysis in the previous section can be used directly. Namely, using \cref{eq:min_norm_sum_equals_sum_norm}, the following equalities hold
\begin{equation}
    \label{eq:norm_sums_equals_sum_norms_views}
    \begin{split}
    & \underset{w \in \mathbb{S}_+^{PL-1}}{\arg\min} \lVert \sum_i^P [\Tilde{W}^{(K)}]_i [\Tilde{w}]_i \rVert_2 \\
    &= \underset{w \in \mathbb{S}_+^{P-1}, \beta \in \mathbb{S}_+^{L-1}}{\arg\min} \lVert \sum_i^P w_i [\Tilde{W}^{(K)}]^i \beta^i \rVert_2 \\
    &= \underset{w \in \mathbb{S}_+^{P-1}, \beta \in \mathbb{S}_+^{L-1}}{\arg\min}  \sum_i^P w_i \lVert [\Tilde{W}^{(K)}]^i \beta^i \rVert_2 \\
    \end{split}
\end{equation}
where $[\Tilde{W}^{(K)}]^i$ denotes the block of columns in $\Tilde{W}^{(K)}$ pertaining to primitive $i$. Similar to \Cref{eq:min_norm_sum_equals_sum_norm} linearizing out $w_i$, $\beta^i$ can be linearized out to produce a view metric that only requires storing and updating per-Gaussian attributes 

\begin{equation}
    \label{eq:view_metric_linearized_beta}
    \mathcal{I}^{(K+1)}_{\text{view}}(x_0, d) =   \sum_i^P w_i(x_0, d) \sum_\ell^L \beta^i_\ell (d) \lVert [\Tilde{W}^{(K)}]^i_\ell \rVert_2,
\end{equation}
which is the view-dependent analogue to \Cref{eq:info_gain_render}. 

Although simple, in practice, we find several reasons for concern for using Equations (\ref{eq:min_norm_sum_equals_sum_norm}), (\ref{eq:info_gain_render}), (\ref{eq:norm_sums_equals_sum_norms_views}), or (\ref{eq:view_metric_linearized_beta}) as a view metric. First, computing the transmittance terms for every pixel-primitive pair in the training data is computationally expensive and memory-hungry. Second, these transmittance values are typically noisy and can change rapidly during the training process as primitives pass in front of each other. This noisy metric induces higher variance between reconstructed models. In addition, we should keep in mind that the 3DGS is a proxy of the ground-truth geometry. Therefore, intertwining the view metric too deeply with the 3DGS parameters leads to suboptimal reconstruction of the ground-truth. We find that abstracting away transmittance effects leads to more reliable behavior, as shown in \Cref{tab:ablation_30k_grad}. 

\begin{figure*}[th]
    \centering
    \includegraphics[width=\textwidth]{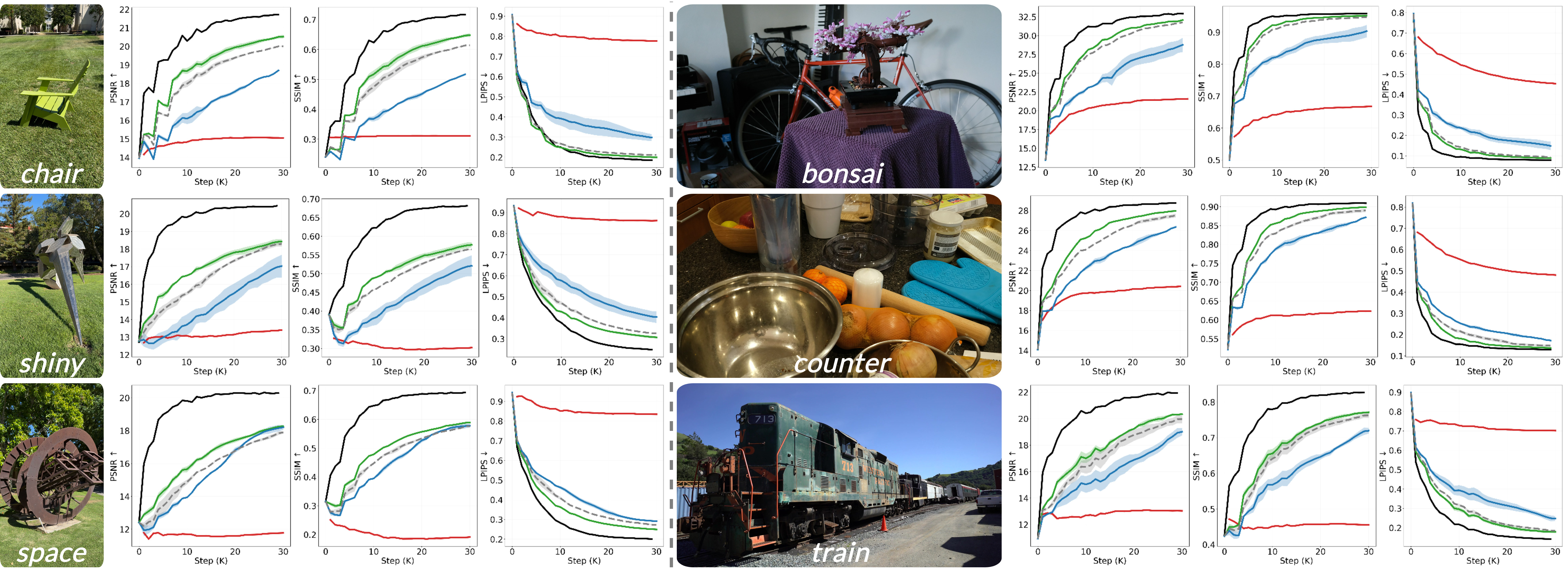}
    \caption{Image-based metrics (PSNR/SSIM/LPIPS) across several scenes for five view selection methods: {\textcolor{red}{Bayes' Rays}}, {\textcolor{blue}{FisherRF}}, {\textcolor{gray}{Random}},
    {\textcolor{green}{COVER}},
    and an infeasible oracle (black) trained with all training images.}
    \label{fig:quantitative}
\end{figure*}
\subsection{Transmittance-Agnostic Metric}
We abstract away the noisy transmittance effects induced by the training cameras by treating all primitives that appear in the frustum of a camera as equal in weight. Primitives outside the frustum are still set to $0$ as usual. As a result, $[\Tilde{W}^{(K)}]^i$ has the following structure

\begin{equation}
    \label{eq:tilde_w_structure}
    [\Tilde{W}^{(K)}]^i = \begin{pmatrix}
        \vdots\\
        A_c\\
        \vdots\\
    \end{pmatrix} \; , \; A_c = 
        \textbf{vis}_c^i  \otimes \beta^i(d_c^i),
\end{equation}
where $\textbf{vis}_c^i \in \{0, 1\}^M$ is the vector of visibilities of primitive $i$ for all $M$ pixels in camera $c$. $d^i_c$ is the viewing direction of camera $c$ incident on primitive $i$. Equivalently,

\begin{equation}
    \label{eq:tilde_w_beta_structure}
    [\Tilde{W}^{(K)}]^i \beta^i_{test} = \begin{pmatrix}
        \vdots\\
        \alpha^i_c \textbf{vis}_c^i \\
        \vdots\\
    \end{pmatrix},
\end{equation}
where $\alpha^i_c = \beta^i(d^i_c) \cdot \beta^i_{test}$ is the dot product between the patch weights associated with training camera $c$ on primitive $i$ with those of the candidate camera on the same primitive. 

The 2-norm of \Cref{eq:tilde_w_beta_structure} is
\begin{equation}
    \label{eq:tilde_w_beta_norm}
    \lVert [\Tilde{W}^{(K)}]^i \beta^i_{test} \rVert_2^2 = \sum_c (\alpha_c^i)^2 |\textbf{vis}_c^i|,
\end{equation}
where $|\textbf{vis}_c^i|$ is the number of pixels in camera $c$ that see primitive $i$. 

We make the simplifying assumption that $|\textbf{vis}_c^i|$ is constant across cameras, allowing us to effectively ignore the contribution of this term in the optimization. In addition, we assume that $\sum_c (\alpha_c^i)^2 \approx \max_c (\alpha_c^i)^2$, which holds true if there is a dominant $\alpha_c^i$ across cameras.

For the purposes of extracting a computable metric, we assume a specific structure to $\beta^i$, namely the spherical Gaussian  kernel, though the derivation can be broadly extended to decaying spherical kernels. A spherical Gaussian kernel has the form
\begin{equation}
    \label{eq:linear_kernel}
    \beta(d; \mu, \kappa) = \mathcal{C} \exp(\kappa d\cdot \mu),
\end{equation}
for normalization constant $\mathcal{C}$, concentration parameter $\kappa$, and mean direction $\mu$.

Therefore, 
\begin{equation}
\label{eq:simple_gaussian_kernel_dot_product}
\begin{split}
    \alpha_c^i &= \beta^i(d^i_c) \cdot \beta^i_{test} \\
    & = \sum_\ell \mathcal{C}^2 \exp(\kappa d_\ell \cdot d^i_c) \cdot \exp(\kappa d_\ell \cdot d^i_{test})\\
    &= \sum_\ell \mathcal{C}^2 \exp(\kappa d_\ell \cdot (d^i_c + d^i_{test}))
\end{split}
\end{equation}
where $d_\ell$ are the directions discretized over the unit sphere.

Note that the choice of $\kappa$ is a design decision and can be selected as an arbitrary value. If $\kappa$ is large, then the color seen from direction $d$ corresponds to the color of the color field patch with direction $d_\ell$ closest to $d$ (a natural choice). As a result, $\alpha^i_c$ is dominated by  the term associated with $d_\ell = (d^i_c + d^i_{test})$. Thus, $\alpha_c^i \approx \mathcal{C}^2 \exp(\kappa \lVert d^i_c + d^i_{test}\rVert_2^2)$. The exponential can be approximated by its first-order Taylor expansion
\begin{equation}
    \label{eq:approximation}
    \begin{split}
    \alpha_c^i &\approx \mathcal{C}^2 (1 + \kappa\lVert d^i_c + d^i_{test}\rVert_2^2) \\
    &= \mathcal{C}^2 (1 + \kappa (2 + 2 d^i_c \cdot d^i_{test})).
    \end{split}
\end{equation}

Combining \cref{eq:approximation} and \cref{eq:tilde_w_beta_norm}, 
\begin{equation}
    \label{eq:tilde_w_beta_norm_2}
    \begin{split}
    \lVert [\Tilde{W}^{(K)}]^i \beta^i_{test} \rVert_2 &\approx \max_c \alpha_c^i \\
    &\propto \frac{1 + \max_c d^i_c \cdot d^i_{test}}{2}.
    \end{split}
\end{equation}
Consequently, our coverage-based, transmittance-agnostic information gain metric is

\begin{equation}
    \label{eq:coverage_render}
    \mathcal{I}^{(K+1)}_{\text{cov}}(x_0, d) = \sum_i^P w_i(x_0, d) \frac{1 + \max_c d^i_c \cdot d}{2},
\end{equation}
which again is renderable and interpretable. Intuitively, the metric favors a camera whose viewing direction is angularly different from all existing training views for the Gaussians it sees, thereby encouraging acquisition of novel geometric coverage. In practice, computing this view metric is efficient. Instead of storing all training view directions per Gaussian, we can store a discretized grid on the unit sphere per Gaussian, with each patch being 0 or 1. Any training camera that was incident on that Gaussian results in the patch boolean corresponding to the direction of that camera to flip to 1. When evaluating the metric, the test view direction is dotted against all unit directions corresponding to the discretized grid, but masked using the grid booleans before taking the max. 

Another advantage of the coverage metric (\ref{eq:coverage_render}) is the ability to naturally bias towards exploration or exploitation as a consequence of its interpretability and boundedness. Similar to the existing alpha compositing of the color render with the background, a background term $b \in \{0,1\}$ can be composited with $\mathcal{I}_{\text{cov}} \in [0, 1]$. Setting $b=1$ rewards foreground occlusion (encouraging exploitation), while $b=0$ penalizes it (favoring exploration). To balance these objectives and avoid querying empty space (e.g. the sky), our implementation uses a hybrid approach: averaging pixels within the non-zero alpha mask and using a $b=0$ background. Because the background saturates the alpha to 1, we render normally without normalizing the weight vector (i.e. $\lVert [w, b] \rVert_1 = 1$ instead of $\lVert [w,b] \rVert_2 = 1$), due to unnecessary added complexity. Additional discussion of the metric can be found in \Cref{app:differentiability}.

\section{Results}
\label{Sec:Results}

We benchmark COVER against state-of-the-art active view planning metrics (Bayes' Rays \cite{goli2024bayes} and FisherRF \cite{jiang2023fisherrf}) in a next-best-view selection task using a fixed dataset (i.e. no arbitrary viewpoints). Additionally, in order to contextualize the gains any method exhibits over any other, we also implement a random baseline that randomly chooses a camera from the candidate set at every time step. Lastly, we  include results from an oracle that has access to all training images at initialization, which is not a feasible policy but simply serves as a loose upper bound on performance. Each method chooses a sequence of camera poses, one-by-one, to be added to the training data set while the $3$D scene representation is actively training. Chosen cameras are removed from the candidate pool of cameras at the next time step. Experiments were run on 3 scenes from the Tanks and Temples~\cite{knapitsch2017tanks} dataset, the entirety of the MipNeRF360 dataset~\cite{barron2022mip}, and 3 custom scenes that were collected using a handheld phone, for a total of 15 scenes. The scene is initially seeded with 10 views in the training dataset. Then, a new view is chosen every 200 gradient steps and all methods are terminated at 30K gradient steps. All view selection metrics and training is implemented within the Nerfstudio framework~\cite{tancik2023nerfstudio}. Each method was run multiple times for reproducibility except for Bayes' Rays due to its slow compute times.

\subsection{Fixed Dataset Photometric Comparisons}

We observe that a random baseline is performant for view selection on a fixed dataset. Visually, random is generally similar in reconstruction quality to COVER (\cref{fig:evalset_viz}). Human-captured datasets are naturally  well-distributed and facilitate  high quality reconstruction. Therefore, random inherits this dataset bias and achieves good coverage. Yet, COVER is still visually and qualitatively superior. For example, there are sometimes artifacts in the random baseline that suggest a lack of coverage (\cref{fig:evalset_viz}), which may not always manifest in large differences in image-based metrics. Overall, COVER outperforms all feasible baselines in typical image-based metrics (i.e. PSNR/SSIM/LPIPS). In \Cref{fig:quantitative}, Bayes' Rays performs the worst, naturally inheriting the performance gap present between $3$DGS and NeRFs. FisherRF demonstrates a significant performance gap with random and COVER. Although there is a smaller gap, we still observe a performance gain of COVER over random (\Cref{tab:summary_30k_structured}). In fact, COVER approaches the infeasible oracle in \texttt{bonsai} and \texttt{counter} despite only observing half of the full dataset. 

\TableSummaryThirtyK

\subsection{Ablations}
Additionally, we ablated two different experimental setups: 1 initial view (Sparse) and an embodied view selection process (Embodied). The embodied process utilizes iterative k-NN ($K=5$) selection to select the best scoring frames on finite datasets to simulate continuous deployment (i.e. no teleporting). This approach allows us to rigorously evaluate performance against ground truth, avoiding the ambiguity of open-ended exploration vs reconstruction metrics. In practice, the finite dataset is useful to implicitly anchor the task, biasing the views toward relevant scene content (i.e. away from walls or floors). COVER performs even better than random in the embodied scenario. With just sparse initialization, random and COVER are identical. However, combining the sparse initialization with the embodied selection scheme broadens the gap to 1.5 PSNR, suggesting the applicability of COVER on robots performing in-the-wild scene reconstruction.
\TableEmbodiedSparseAblationThirtyK

\subsection{Compute Time}
COVER is as efficient as it is performant. Our method requires on average $3.5$ seconds to sweep through the whole dataset ($>300$ images) to choose the best view. Note that COVER does not utilize any additional custom CUDA kernels beyond what is available in Nerfstudio and the gsplat library \cite{ye2025gsplat}. Meanwhile, FisherRF requires on average $23.9$ seconds, likely due to the computation of gradient information. Finally, Bayes' Rays requires on average $37.1$ seconds. In fixed time budget settings (e.g. on a real-time robot system), COVER is appealing as it can process many more images than other methods, resulting in better optimality of the chosen view. 

\section{Limitations}
\label{Sec:Limitations}

COVER is derived from a set of approximations that trade off fidelity for scalability. In particular, the metric relies on a coverage-based surrogate that lower-bounds the Fisher Information Gain while discarding explicit transmittance effects. Although effective and highly efficient, this approximation may be less reliable in scenes with extreme clutter where transmittance carries additional information, though we have not observed this behavior in commonly used datasets. Additionally, COVER is illumination-agnostic: it selects views based solely on geometric coverage and accumulated visibility, without explicitly modeling shading, lighting direction, or photometric variation. As a result, its selections may be suboptimal for tasks where appearance changes dominate, such as scenes with time-varying illumination or materials with complex BRDFs. The method also assumes access to a reasonably accurate intermediate reconstruction from which per-primitive visibility can be estimated. As shown in our results, our method is only as good as the random baseline in very sparse initialization regimes where early inaccuracies in geometry or Gaussian placement may affect ranking quality. Finally, we simulate embodied execution in our results. Extending COVER to online planning with robot-constrained trajectories on real hardware and across a broader range of radiance-field backbones is a promising direction for future work.

\section{Conclusion}
\label{Sec:Conclusion}

COVER is a simple and efficient next-best-view metric grounded in an analysis of the Fisher Information for radiance fields. Our derivation shows that geometric coverage emerges as a dominant factor controlling the information contributed by a new observation. This insight leads to a practical view-selection criterion that avoids explicit transmittance estimation, integrates cleanly into existing radiance field pipelines, and can be evaluated and visualized in real time.

Across a variety of real-world scenes and between fixed and embodied data acquisition schemes, COVER consistently improves reconstruction quality relative to random and Fisher-information-based baselines, while adding negligible overhead and requiring no model modifications. The ability to compute the metric directly from intermediate training states further enhances its usability for incremental datasets and active acquisition.

Overall, our results highlight that a principled yet lightweight coverage formulation can serve as an effective proxy for information gain in radiance-field reconstruction. Future work will explore extensions to online active mapping, trajectory-aware selection, and illumination-aware or task-specific variants of the coverage metric.

\section{Acknowledgments}
This work is supported in part by ONR grant N00014-23-1-2354.  The first author is supported on a NASA NSTGRO fellowship, the second author is supported by Blue Origin, and the third author is supported on a NDSEG fellowship.  We are grateful for this support.

{
    \small
    \bibliographystyle{ieeenat_fullname}
    \bibliography{main}

\begin{thebibliography}{23}
\providecommand{\natexlab}[1]{#1}
\providecommand{\url}[1]{\texttt{#1}}
\expandafter\ifx\csname urlstyle\endcsname\relax
  \providecommand{\doi}[1]{doi: #1}\else
  \providecommand{\doi}{doi: \begingroup \urlstyle{rm}\Url}\fi

\bibitem[Barron et~al.(2022)Barron, Mildenhall, Verbin, Srinivasan, and Hedman]{barron2022mip}
Jonathan~T Barron, Ben Mildenhall, Dor Verbin, Pratul~P Srinivasan, and Peter Hedman.
\newblock Mip-nerf 360: Unbounded anti-aliased neural radiance fields.
\newblock In \emph{Proceedings of the IEEE/CVF conference on computer vision and pattern recognition}, pages 5470--5479, 2022.

\bibitem[Goli et~al.(2024)Goli, Reading, Sell{\'a}n, Jacobson, and Tagliasacchi]{goli2024bayes}
Lily Goli, Cody Reading, Silvia Sell{\'a}n, Alec Jacobson, and Andrea Tagliasacchi.
\newblock {Bayes' Rays: Uncertainty Quantification for Neural Radiance Fields}.
\newblock In \emph{Proceedings of the IEEE/CVF Conference on Computer Vision and Pattern Recognition}, pages 20061--20070, 2024.

\bibitem[Jiang et~al.(2023)Jiang, Lei, and Daniilidis]{jiang2023fisherrf}
Wen Jiang, Boshu Lei, and Kostas Daniilidis.
\newblock {FisherRF: Active View Selection and Uncertainty Quantification for Radiance Fields using Fisher Information}.
\newblock \emph{arXiv preprint arXiv:2311.17874}, 2023.

\bibitem[Kerbl et~al.(2023)Kerbl, Kopanas, Leimkuehler, and Drettakis]{kerbl2023splatting}
Bernhard Kerbl, Georgios Kopanas, Thomas Leimkuehler, and George Drettakis.
\newblock 3d gaussian splatting for real-time radiance field rendering.
\newblock \emph{ACM Trans. Graph.}, 42\penalty0 (4), 2023.

\bibitem[Knapitsch et~al.(2017)Knapitsch, Park, Zhou, and Koltun]{knapitsch2017tanks}
Arno Knapitsch, Jaesik Park, Qian-Yi Zhou, and Vladlen Koltun.
\newblock Tanks and temples: Benchmarking large-scale scene reconstruction.
\newblock \emph{ACM Transactions on Graphics (ToG)}, 36\penalty0 (4):\penalty0 1--13, 2017.

\bibitem[Li et~al.(2025)Li, Kuang, Li, Hao, Yan, Zhou, and Zhang]{li2025activesplat}
Yuetao Li, Zijia Kuang, Ting Li, Qun Hao, Zike Yan, Guyue Zhou, and Shaohui Zhang.
\newblock Activesplat: High-fidelity scene reconstruction through active gaussian splatting.
\newblock \emph{IEEE Robotics and Automation Letters}, 2025.

\bibitem[Lin and Yi(2022)]{lin2022active}
Kevin Lin and Brent Yi.
\newblock Active view planning for radiance fields.
\newblock In \emph{Robotics Science and Systems}, 2022.

\bibitem[Max(1995)]{max1995opticalmodel}
N. Max.
\newblock Optical models for direct volume rendering.
\newblock \emph{IEEE Transactions on Visualization and Computer Graphics}, 1\penalty0 (2):\penalty0 99--108, 1995.

\bibitem[Mildenhall et~al.(2021)Mildenhall, Srinivasan, Tancik, Barron, Ramamoorthi, and Ng]{mildenhall2021nerf}
Ben Mildenhall, Pratul~P. Srinivasan, Matthew Tancik, Jonathan~T. Barron, Ravi Ramamoorthi, and Ren Ng.
\newblock Nerf: representing scenes as neural radiance fields for view synthesis.
\newblock \emph{Commun. ACM}, 65\penalty0 (1):\penalty0 99–106, 2021.

\bibitem[Nagami et~al.(2025)Nagami, Chen, Yu, Shorinwa, Adang, Dougherty, Cristofalo, and Schwager]{nagami2025vista}
Keiko Nagami, Timothy Chen, Javier Yu, Ola Shorinwa, Maximilian Adang, Carlyn Dougherty, Eric Cristofalo, and Mac Schwager.
\newblock {VISTA: Open-Vocabulary, Task-Relevant Robot Exploration with Online Semantic Gaussian Splatting}.
\newblock \emph{arXiv preprint arXiv:2507.01125}, 2025.

\bibitem[Pan et~al.(2022)Pan, Lai, Song, and Huang]{pan2022activenerf}
Xuran Pan, Zihang Lai, Shiji Song, and Gao Huang.
\newblock Activenerf: Learning where to see with uncertainty estimation.
\newblock In \emph{European Conference on Computer Vision}, pages 230--246. Springer, 2022.

\bibitem[Shen et~al.(2023)Shen, Yang, Yu, Wong, Kaelbling, and Isola]{shen2023F3RM}
William Shen, Ge Yang, Alan Yu, Jansen Wong, Leslie~Pack Kaelbling, and Phillip Isola.
\newblock Distilled feature fields enable few-shot language-guided manipulation.
\newblock In \emph{7th Annual Conference on Robot Learning}, 2023.

\bibitem[Shorinwa et~al.(2024)Shorinwa, Tucker, Smith, Swann, Chen, Firoozi, Kennedy, and Schwager]{shorinwa2024splat}
Ola Shorinwa, Johnathan Tucker, Aliyah Smith, Aiden Swann, Timothy Chen, Roya Firoozi, Monroe~David Kennedy, and Mac Schwager.
\newblock Splat-mover: Multi-stage, open-vocabulary robotic manipulation via editable gaussian splatting.
\newblock 2024.

\bibitem[Strong et~al.(2025)Strong, Lei, Swann, Jiang, Daniilidis, and Kennedy]{strong2025next}
Matthew Strong, Boshu Lei, Aiden Swann, Wen Jiang, Kostas Daniilidis, and Monroe Kennedy.
\newblock Next best sense: Guiding vision and touch with fisherrf for 3d gaussian splatting.
\newblock In \emph{2025 IEEE International Conference on Robotics and Automation (ICRA)}, pages 3204--3210. IEEE, 2025.

\bibitem[Tancik et~al.(2023)Tancik, Weber, Ng, Li, Yi, Wang, Kristoffersen, Austin, Salahi, Ahuja, et~al.]{tancik2023nerfstudio}
Matthew Tancik, Ethan Weber, Evonne Ng, Ruilong Li, Brent Yi, Terrance Wang, Alexander Kristoffersen, Jake Austin, Kamyar Salahi, Abhik Ahuja, et~al.
\newblock Nerfstudio: A modular framework for neural radiance field development.
\newblock In \emph{ACM SIGGRAPH 2023 conference proceedings}, pages 1--12, 2023.

\bibitem[Tao et~al.(2025)Tao, Ong, Murali, Spasojevic, Chaudhari, and Kumar]{tao2025rt}
Yuezhan Tao, Dexter Ong, Varun Murali, Igor Spasojevic, Pratik Chaudhari, and Vijay Kumar.
\newblock Rt-guide: Real-time gaussian splatting for information-driven exploration.
\newblock \emph{IEEE Robotics and Automation Letters}, 2025.

\bibitem[Williams and Max(1992)]{max1992dvolumedensity}
Peter~L. Williams and Nelson Max.
\newblock A volume density optical model.
\newblock In \emph{Proceedings of the 1992 Workshop on Volume Visualization}, page 61–68, New York, NY, USA, 1992. Association for Computing Machinery.

\bibitem[Xiao et~al.(2024)Xiao, Santa~Cruz, Ahmedt-Aristizabal, Salvado, Fookes, and Lebrat]{xiao2024nerf}
Wenhui Xiao, Rodrigo Santa~Cruz, David Ahmedt-Aristizabal, Olivier Salvado, Clinton Fookes, and L{\'e}o Lebrat.
\newblock Nerf director: Revisiting view selection in neural volume rendering.
\newblock In \emph{CVPR}, 2024.

\bibitem[Xie et~al.(2023)Xie, Zhang, Li, Zhang, and Zhang]{xie2023s}
Ziyang Xie, Junge Zhang, Wenye Li, Feihu Zhang, and Li Zhang.
\newblock S-nerf: Neural radiance fields for street views.
\newblock \emph{arXiv preprint arXiv:2303.00749}, 2023.

\bibitem[Xu et~al.(2025)Xu, Jin, Wu, Zhao, Zhang, Zhao, Gao, Gan, and Ding]{xu2025hgs}
Zijun Xu, Rui Jin, Ke Wu, Yi Zhao, Zhiwei Zhang, Jieru Zhao, Fei Gao, Zhongxue Gan, and Wenchao Ding.
\newblock Hgs-planner: Hierarchical planning framework for active scene reconstruction using 3d gaussian splatting.
\newblock In \emph{2025 IEEE International Conference on Robotics and Automation (ICRA)}, pages 14161--14167. IEEE, 2025.

\bibitem[Xue et~al.(2024)Xue, Dill, Mathur, Dellaert, Tsiotras, and Xu]{xue2024neural}
Shangjie Xue, Jesse Dill, Pranay Mathur, Frank Dellaert, Panagiotis Tsiotras, and Danfei Xu.
\newblock Neural visibility field for uncertainty-driven active mapping.
\newblock In \emph{CVPR}, 2024.

\bibitem[Yan et~al.(2023)Yan, Liu, Quan, Chen, and Fu]{yan2023active_implicit}
Dongyu Yan, Jianheng Liu, Fengyu Quan, Haoyao Chen, and Mengmeng Fu.
\newblock Active implicit object reconstruction using uncertainty-guided next-best-view optimization.
\newblock \emph{IEEE Robotics and Automation Letters}, 2023.

\bibitem[Ye et~al.(2025)Ye, Li, Kerr, Turkulainen, Yi, Pan, Seiskari, Ye, Hu, Tancik, and Kanazawa]{ye2025gsplat}
Vickie Ye, Ruilong Li, Justin Kerr, Matias Turkulainen, Brent Yi, Zhuoyang Pan, Otto Seiskari, Jianbo Ye, Jeffrey Hu, Matthew Tancik, and Angjoo Kanazawa.
\newblock gsplat: An open-source library for gaussian splatting.
\newblock \emph{Journal of Machine Learning Research}, 26\penalty0 (34):\penalty0 1--17, 2025.

\end{thebibliography}
}

\appendix
\clearpage

\section{Proofs}
\label{app:proof1}

The RHS (\ref{eq:min_norm_sum_equals_sum_norm}) is simply a linear objective of the form $\sum_i \alpha_i w_i$. The constraint set $w \in \mathbb{S}^{P-1}_+$ implies that $\lVert w \rVert_1 \geq 1$ and consequently, $\sum_i w_i \geq 1$. Therefore, the following inequalities hold 

\begin{equation}
    \label{eq:minimizer_linear_inequalities}
    \sum_i \alpha_i w_i \geq (\min_i \alpha_i ) \sum_i w_i \geq \min_i \alpha_i,
\end{equation}
with equality when $a = e_{i^*}$, the one-hot vector corresponding to $i^* = \arg\min_i \alpha_i$. Any mixture is necessarily bigger by triangle inequality and because the problem is convex, $e_{i^*}$ is a \emph{unique} minimizer corresponding to the column of $W^{(K)}$ with the \emph{smallest} norm. 

The LHS (\ref{eq:min_norm_sum_equals_sum_norm}) is identical to a quadratic objective, which we show has the same minimizer. Due to the non-negativity of $W$, dot products between columns of $W$ must be non-negative. Subsequent inequalities follow
\begin{equation}
\label{eq:minimizer_quadratic_inequalities}
    \begin{split}
    w^T G w &= \sum_i w_i^2 \lVert W_{:, i} \rVert_2^2 + 2 \sum_{i < j} w_i w_j W_{:, i}^T W_{:, j}\\
    & \geq \sum_i w_i^2 \lVert W_{:, i} \rVert_2^2\\
    &\geq (\min_i \lVert W_{:, i} \rVert_2^2) \sum_i w_i^2 = \min_i \lVert W_{:, i} \rVert_2^2,
    \end{split}
\end{equation}
with equality when $a = e_{i^*}$, where $i^* = \arg\min_i  \lVert W_{:, i} \rVert_2$ corresponds to the \emph{smallest} norm column. With the one-hot vector, the cross terms $w_i w_j$ are 0 if $i \neq j$, and the latter inequalities follow automatically. As a result, the \emph{unique} minimizer of the LHS of \cref{eq:min_norm_sum_equals_sum_norm} is also a minimizer of the RHS (\ref{eq:min_norm_sum_equals_sum_norm}).

\section{Additional Results}
\label{sec:appendix_tables}
We observe that most methods display a logarithmic-shaped curve for image-based metrics over time. Based on this observation, we find that comparing methods very early or very late in training was not very informative. Too early and the methods have not had sufficient time to accumulate errors; too late and the methods will have chosen many of the same views, leading to similar reconstruction quality. In addition, neither are common situations, as the view selection seeks to retrieve the highest fidelity scene representation as efficiently as possible.

Therefore, we quantitatively compare all methods by showing image-based metrics at $10$K (\cref{fig:tnt_comparison_10K,fig:custom_comparison_10K,fig:mipnerf360_comparison_10K}) and $30$K (\cref{fig:tnt_comparison_30K,fig:custom_comparison_30K,fig:mipnerf360_comparison_30K}) gradient steps. In addition, we report an Area-Under-Curve (AUC) metric (normalized by the number of gradient steps), the integrated difference between a method and random over time. We denote this metric with a $\Delta$.  The AUC metric indicates the time-averaged improvement of the method over random. We observe that COVER consistently outperforms the competing methods regardless of the sweep over time.

\TableTNTTenK
\TableCapturesTenK
\TableMipNeRFTenK

\TableTNTThirtyK
\TableCapturesThirtyK
\TableMipNeRFThirtyK

\section{Optimization in Continuous Space and Transmittance-based Metrics}
\label{app:differentiability}
Additionally, we investigate the viability of transmittance-based view metrics and the use of COVER in a continuous space optimization scheme. To test, we compare \Cref{eq:view_metric_linearized_beta} and COVER against an extension (Gradient), which initializes around the best neighbor (root) and then performs 50 gradient descent steps into the camera pose to minimize coverage, showcasing free-view selection and continuous optimization. Since our datasets are finite, we instead take renders from a trained Splatfacto model at these optimized poses and add them to the training set, in addition to the root's frame to stabilize training.

\TableEmbodiedAblationThirtyK

COVER is superior to \Cref{eq:view_metric_linearized_beta}, suggesting that designing the view metric to accommodate transmittance effects when they can be noisy and transient degrades optimality of the chosen viewpoint.  COVER outperforms Gradient because Gradient's training images are partially derived from a 3DGS rather than reality. Note that not including the root frames (which came from ground-truth images) in the training set destabilizes training, which suggests that training a 3DGS using auxiliary 3DGS renders can lead to negative feedback. We believe with access to a simulator or execution on hardware, Gradient might slightly outperform COVER.

\section{Additional Visualizations}
\begin{figure*}
    \centering
    \includegraphics[width=\linewidth]{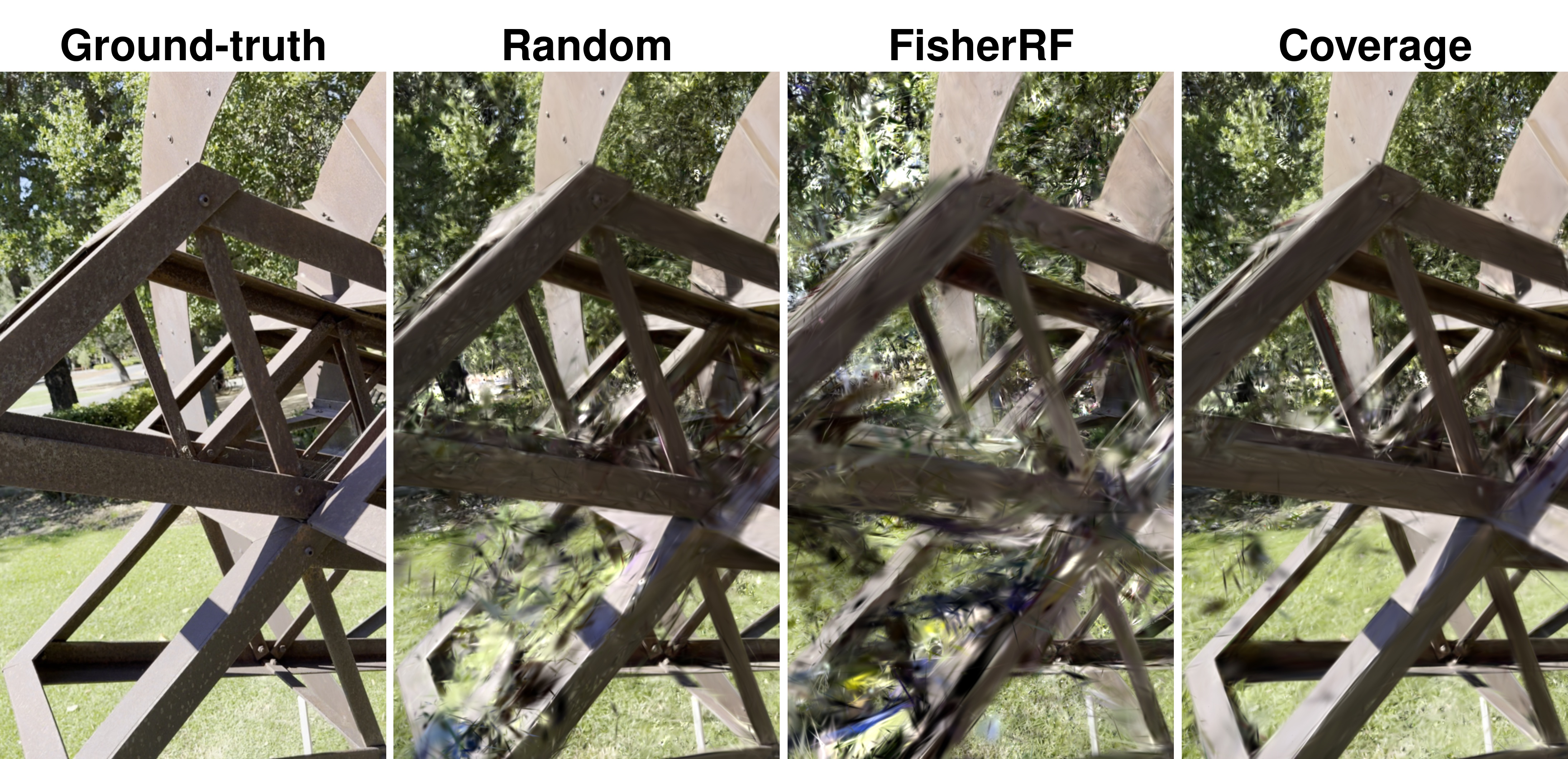}
    \caption{Comparison of different view metrics in the \texttt{space} scene.}
    \label{fig:space_coverage}
\end{figure*}

\begin{figure*}
    \centering
    \includegraphics[width=\linewidth]{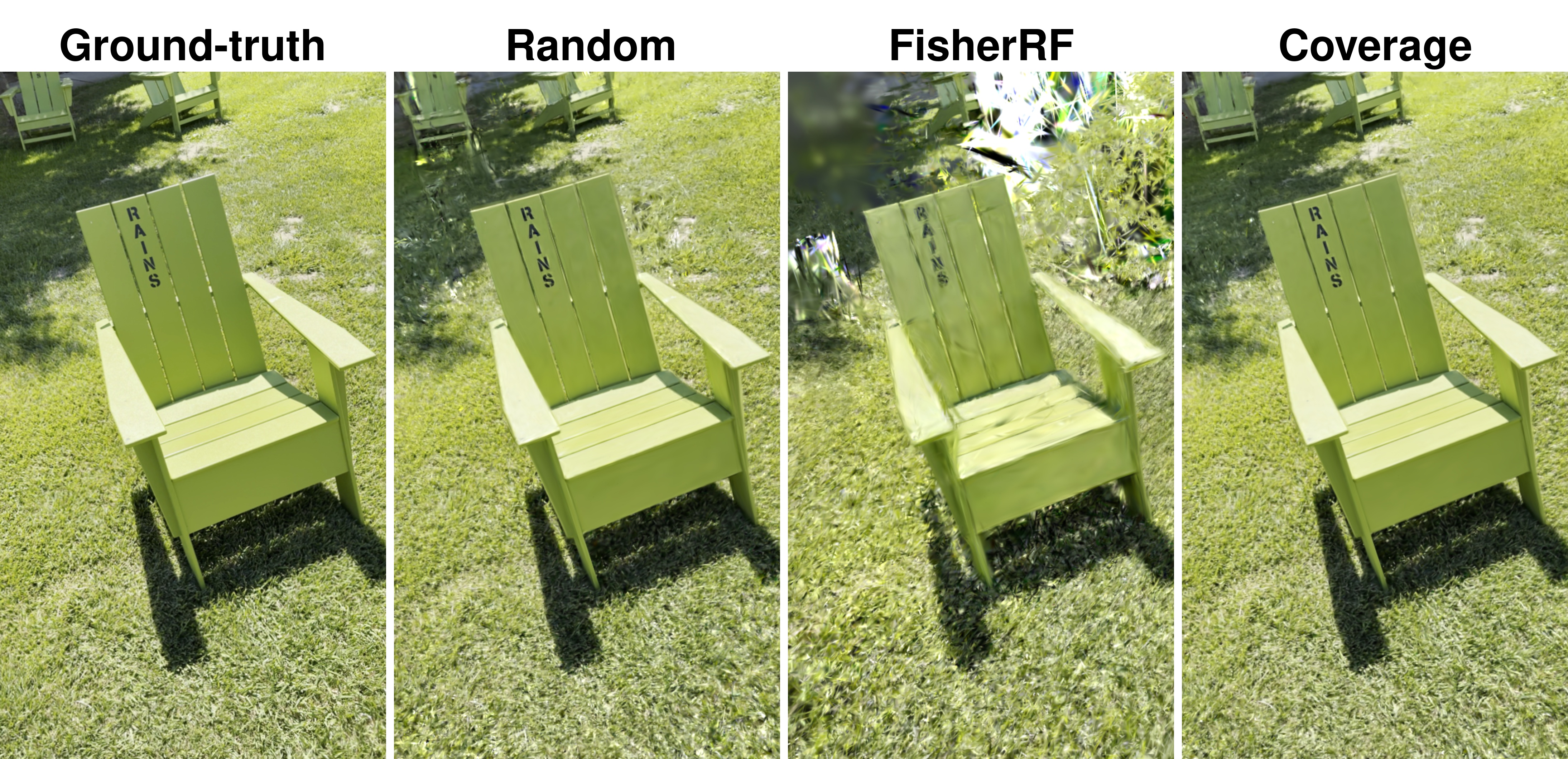}
    \caption{Comparison of different view metrics in the \texttt{chair} scene.}
    \label{fig:chair_coverage}
\end{figure*}

We present more visual comparisons between the quality of COVER and the baselines in \Cref{fig:space_coverage} and \Cref{fig:chair_coverage}. We observe better reconstruction for both of these scenes. We encourage readers to zoom in to notice the details.

\label{app:more_viz}

\end{document}